\documentclass[conference]{IEEEtran}
\ifCLASSINFOpdf
  \usepackage[pdftex]{graphicx}
\else
\fi
\usepackage[tight,footnotesize]{subfigure}
\usepackage{subfloat}
\usepackage[caption=true]{caption}

\begin{document}
%
\title{A hypothesize-and-verify framework for Text Recognition using Deep Recurrent Neural Networks}
\author{\IEEEauthorblockN{Anupama Ray}
\IEEEauthorblockA{Department of Electrical Engineering\\
Indian Institute of Technology Delhi\\
Email: anupamaray88@gmail.com}
\and
\IEEEauthorblockN{Sai Rajeswar}
\IEEEauthorblockA{Department of Electrical Engineering\\
Indian Institute of Technology Delhi\\
Email: rajsai24@gmail.com}
\and
\IEEEauthorblockN{Santanu Chaudhury}
\IEEEauthorblockA{Department of Electrical Engineering\\
Indian Institute of Technology Delhi\\
Email: schaudhury@gmail.com}
}


%
\maketitle
\begin{abstract}
Deep LSTM is an ideal candidate for text recognition. However text recognition involves some initial image processing steps like segmentation of lines and words which can induce error to the recognition system. Without segmentation, learning very long range context is difficult and becomes computationally intractable. Therefore, alternative soft decisions are needed at the pre-processing level. This paper proposes a hybrid text recognizer using a deep recurrent neural network with multiple layers of abstraction and long range context along with a language model to verify the performance of the deep neural network. In this paper we construct a multi-hypotheses tree architecture with candidate segments of line sequences from different segmentation algorithms at its different branches. The deep neural network is trained on perfectly segmented data and tests each of the candidate segments, generating unicode sequences. In the verification step, these unicode sequences are validated using a sub-string match with the language model and best first search is used to find the best possible combination of alternative hypothesis from the tree structure. Thus the verification framework using language models eliminates wrong segmentation outputs and filters recognition errors. 
\end{abstract}
%
\IEEEpeerreviewmaketitle

\section{Introduction}

Most Optical Character Recognition (OCR) algorithms assume perfect segmentation of lines and words, which is not true. In Indic scripts, the presence of vowel modifiers and conjucts furthur aggrevate the errors in segmentation as these modifiers are present in the upper or lower zone. This makes the text layout dense and decreases the interline separation. This paper proposes a text recognition framework to hypothesize and verify the sequences obtained from multiple segmentation techniques using a deep BLSTM network and a language model to verify the performance of the deep neural network. In this paper we aim to find the best possible recognition of word sequences by searching sub-strings of words derived from multiple segmentation routines. We construct a hypothesize-and-verify framework in which candidate segments of word sequences derived from multiple segmentation routines are at different branches. A deep recurrent neural network is trained on perfectly segmented data and tests each of the candidate segments, generating unicode sequences. This work is an extension of the work on printed text recognition using Deep BLSTM wherein Deep BLSTM architecture for text recognition was proposed \cite{my3rd}. In the verification stage these unicode sequences are validated using a sub-string match with the language model and best first search is used to find the best possible combination of alternative hypothesis from the tree structure. The search region uses a spatial context considering the preceeding and suceeding word to find the best match. This algorithm is able to learn the sequence alignment, solving the unicode re-ordering issues. This verification framework eliminates insertion and deletion errors of the recognizer due to the sub-string match with the n-grams. This is a segmentation free script independent framework and in this paper we presents results on Oriya printed text. The language model is independently learnt on the script under recognition and character n-grams are saved. Oriya script is used due to the unavailability of OCR for this script and due to the challenges involved such as the huge number of classes and shape complexities of the script. 

The paper is organized as follows: Section 2 gives a brief review of the work done in this area, Section 3 presents the Deep BLSTM architecture in detail followed by Section 4 where the data processing and multi-hypotheses framework is discussed. The experimental results are presented in Section 5 followed by conclusion in Section 6.

\begin{figure*}
\centering
\includegraphics[scale=0.5]{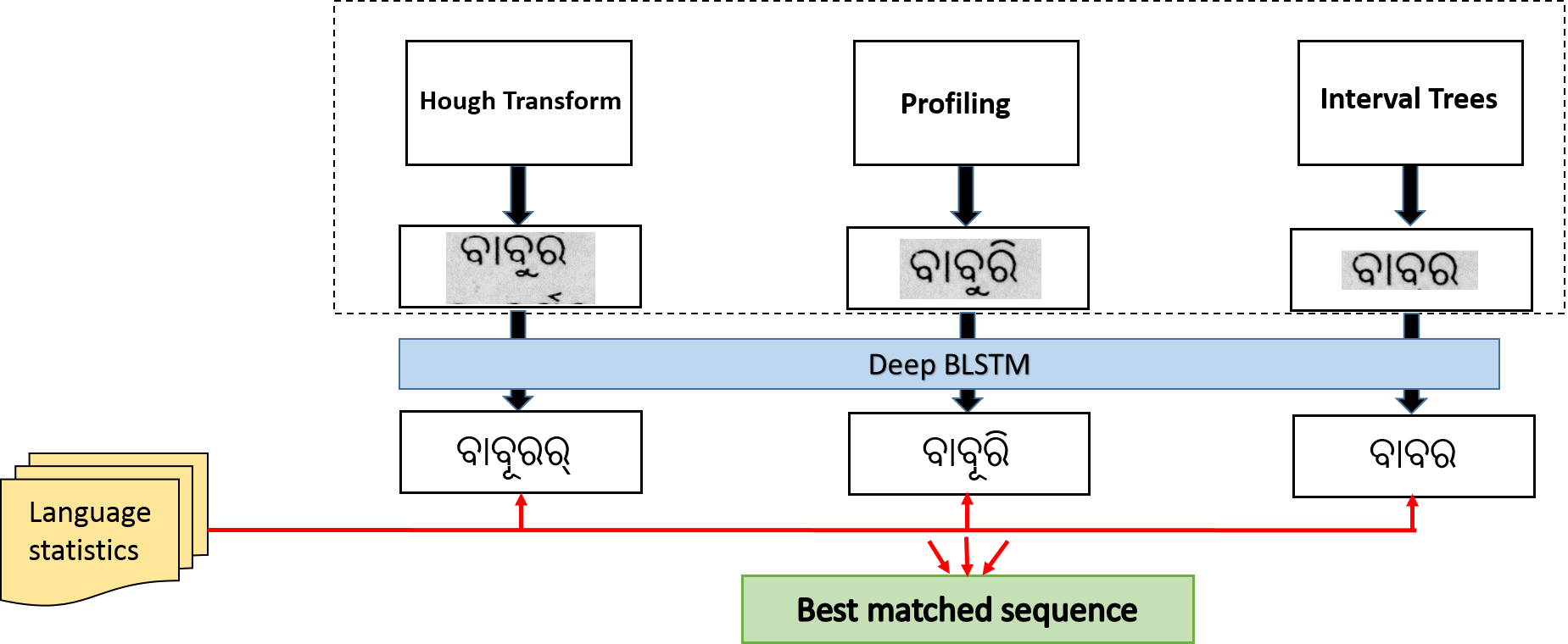}
\caption{Block diagram of Recognition Architecture}
\label{fig1}
\end{figure*}

 

\section{Related Work}

Text recognition algorithms have traditionally been segmentation based where lines are segmented to words and finally characters which get recognized by the use of classifiers. Such approaches have high segmentation error and do not use context information. The main causes of such errors arise from age and quality of documents where inter-word and inter line spacing, ink spread and background text interference cause segmentation errors in turn affecting overall recognition accuracies. In segmentation free approaches sequential classifiers like Hidden Markov Model(HMM) and graphical models like Conditional Random Fields(CRF) have been used. These algorithms introduced the use of context information in terms of transition probabilities and n-gram models, thus improving recognition accuracies\cite{Plamondon:2000:OOH:331097.331275}. But these approaches mostly do not work with unsegmented words, if some do, they are restricted since they use a dictionary of limited words. 

Long Short Term Memory based Recurrent Neural network architecture has been widely used for speech recognition \cite{graves2004biologically, graves2005framewise}, text recognition \cite{liwicki2007novel}, social signal prediction \cite{6854518}, emotion recognition \cite{wollmer2010context} and time series prediction problems since it has the ability of sequence learning. LSTM has emerged as the most competent classifier for handwriting and speech recognition. It performs considerably well on handwritten text without explicit knowledge of the language and has won several competitions \cite{GravesS08, NIPS2007_3213}. LSTM has been used for the recognition of printed Urdu Nastaleeq script \cite{Ul-Hasan:2013:OPU:2549400.2549465} and printed English and Fraktur scripts \cite{Breuel:2013:HOP:2549400.2549524}. RNN based approaches have been popularly used for Arabic scripts wherein segmentation is immensely difficult \cite{Rashid:2013}. LSTM based approaches have outperformed HMM based ones for handwriting recognition proving that learnt features are better than handcrafted features \cite{834}. With the advent of Deep learning algorithms, deep belief networks and deep neural networks are gaining popularity due to their efficiency over shallow models \cite{hinton2006fast}. 

OCRs for Indic scripts are not as robust as that of Roman scripts since most of the algorithms used for Indic script recognition are segmentation based and script dependent \cite{venu:09}. Recognition of Indic script becomes challenging due to various problems as stated here. The nature of Indic scripts is very complex giving rise to huge number of symbols (classes) including basic characters, vowel modifiers, conjuncts formed out of two or more character combination. If the text is noisy or the document is degraded, the recognition suffers badly due to segmentation faults at line and word level. Traditionally, different handcrafted features have mostly been used for text recognition of Oriya \cite{953897}, Bangla\cite{DBLP:conf/icfhr/FinkVBPC10} and classifiers like HMM \cite{DBLP:conf/icpr/ParuiGBC08}, SVM and CRF has been widely used. Naveen et al presented a direct implementation of single layer LSTM network for the recognition of Devanagiri scripts \cite{NaveenICPR, NaveenICDAR} and further experimented on more Indic scripts \cite{NaveenDAS}. 


\section{Recognition Architecture}
For document image recognition we need to segment the full image in order to localize text blocks, lines and words. This segmentation of lines and words from a page induces a lot of error depending upon the age and quality of the document, scanning technique and several such reasons. But a completely segmentation free approach is difficult because learning  very long range sequences can become computationally intractable. Thus at the pre-processing level we incorporate several such segmentation algorithms and use a soft decision based multi-hypothesis architecture for choosing the best possible recognized sequence. In this work we have used three standard segmentation algorithms: Hough Transform, Geometric projection and Interval tree based segmentation \cite{Plamondon:2000:OOH:331097.331275}. Candidate word segments from each segmentation algorithm is passed through the Deep BLSTM recognizer which has been trained on perfect word sequences. The Deep BLSTM network generates output sequences corresponding to the segments. In the multi-hypotheses framework we have same line segments of a page derived out of different segmentation schemes in the different branches and then these sequences are matched using a language model to refine and pick the best sequence. A block diagram of the multi-hypothesis architecture is shown in figure 1.

The main motivation for this hypothesize-and-verify framework comes from the fact that in a test case where we have erroneous segmentation, how do we make the best use of the different segmentation algorithms and the rules of the script to improve recognition. We know that words do not combine in a random order and necessarily follow grammar of the particular script. The order of words determine the grammar and can be learnt from an ideal context model. Since character n-grams are better primitives and have been widely used for retrieval purposes we have used character n-grams instead of word n-grams. Creating a n-gram model of words is difficult as it requires a dictionary of all possible words of the script, which is not available for most Indic scripts. 
The language model is learnt separately for the script to introduce language statistics for rejecting the invalid n-grams and picking the best possible output sequence. Each primitive be it basic character, vowel modifier or conjunct follow a certain order to form a valid word. Here we use trigram and 4-gram and parse the sequence to find the corresponding  matches.

During the sub-string search, if there exists a substring which is not present in the trigram and 4-gram, the substring is considered to be invalid. A cumulative matching score is defined along with a penalty in mismatch with trigram or 4-gram sequence. During faulty line segmentation the upper zone or lower zone primitives (mostly vowel modifiers) combine or miss the original line, thus creating errors. Errorneous word segmentation can also lead to broken characters or parts of word missing.  Such words have faulty unicode sequence due to misplacement or addition of upper or lower zone primitives and thus get misrecongized. These can be taken care of by learning valid combination of characters and we select the best possible sequence out of the different pathways of the multi-hypothesis pipeline. These sequences are verified by using language statistics of a script to find the best possible word. This verification on a multi-hypothesis framework using language models eliminates segmentation errors and is main contribution of this work.

\section{Learning framework}

\subsection{Recurrent Neural Networks}
Deep Recurrent Neural networks (RNNs) have emerged as the very competant classifier for text and speech recognition and Long Short Term Memory has been the most successful recurrent neural network architecture. Bidirectional Long Short Term Memory (BLSTM) has the capability to capture long range context and has succesfully overcome the limitations of standard RNNs like vanishing gradient and need of pre-segmented data. LSTM uses multiplicative gates to trap the error so that a constant error flow is maintained. This phenomenon is called \textit{Constant Error Caraousal} and helps overcome the vanishing gradient problem. Bidirectional LSTM enables accessing longer range context in both directions using forward and backward layers \cite{Graves:2009:NCS:1525650.1525782}.  Graves etal \cite{Graves:2006:CTC:1143844.1143891} proposed a training method known as \textit{Connectionist Temporal Classification} that could align sequential data and thus avoided the need of pre-segmented data. LSTM has emerged as a very successful architecture and is being widely used as a robust OCR architecture for printed and handwritten text \cite{Hochreiter:1997}. Deep networks have outperformed single layer LSTM for speech recognition \cite{DBLP:conf/asru/GravesJM13, DBLP:journals/corr/abs-1303-5778} motivating the use of Deep LSTM architectures for text recognition.

\subsection{Deep BLSTM}
Deep feedforward neural networks refers to having multiple non-linear layers between the input and output layer. But in case of LSTM which is a recurrent neural network, the same principle cannot be applied directly due the temporal structure of RNNs. We construct a deep BLSTM architecture by stacking multiple hidden layers to increase the representational capability of higher order features. RNNs add temporal context to the learning and LSTM's internal cell architecture with the forget gate preserves the state over time. The implementation of deep LSTM with N layers is as follows. This architecture primarily has three bidirectional LSTM layers(BLSTM) used as the three hidden layers(N=3) stacked between the input(N=0) and output(N+1th) layers. 

\begin{equation}
h_{t}^{0} = x_t
\end{equation}
\begin{equation}
h_{t}^{n} = L_{t}^{n}{(h_{t}^{n-1}, h_{t-1}^{n}}
\end{equation}
\begin{equation}
y_t = S{(W^{(N),(N+1)}h_{t}^{N} + b^{N+1})}
\end{equation}

where all superscripts indicate the index of the layer and subscript $t$ denotes the time frame. W is weight matrix, b is the bias, $h_{t}^{n}$ is the hidden layer activation of each memory cell at time t of nth unit (n =1,...N). $L_{t}^{n}$ denotes the activation function of the LSTM. Bidirectional LSTM has been used so that previous and future context with respect to current position can be exploited for sequence learning in both the forward and backward direction in two layers. To create a deep BLSTM network the interlayer connections should be made such that the output of each hidden layer (consisting of a forward and backward LSTM layer) will propagate to both the forward and backward LSTM layer forming the succesive hidden layer. The stacking of hidden layers helps obtain higher level feature abstraction. We have used 36K words for training and 10K for testing. For speedups in the training procedure we harness the power of multicore CPUs by redesigning LSTM as a threaded implementation using OpenMP and BLAS routines.

\begin{figure}[h]
\centering
\includegraphics[scale=0.5]{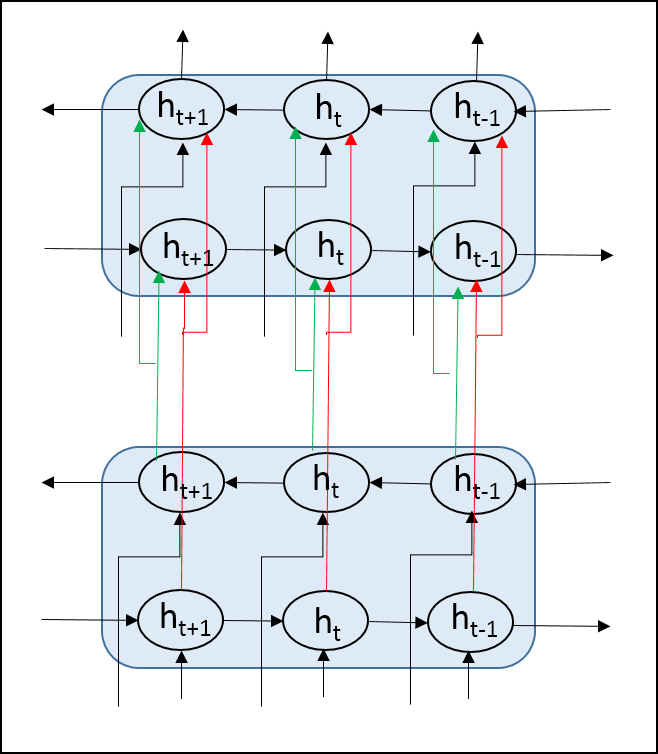}
\centering
\caption{Block Diagram of Deep BLSTM architecture}
\label{fig 2}
\end{figure} 

\subsection{Network Parameters}
The neural network uses CTC output layer with 162 units (161 basic class labels and one for blank). The network is trained with three hidden bidirectional LSTM layers separated by feedforward units with tanh activation. Several experiments have been performed by varying the number of hidden units in each hidden layer. The feedforward layers have tanh activation function and the CTC output layer has softmax activation function. The network is trained with a fixed learning rate of $10^{-4}$, momentum 0.9 and initial weights are selected randomly from [-0.1,0.1]. The total number of weights in the network are 154135. Bias weights to read, write and forget gates are initialized with 1.0, 2.0, -1.0. The output unit squashing function is a sigmoid function. CTC error has been used as the loss function for early stopping since it tends to converge the fastest thereby training time decreases with decrease in the number of epochs. For BLSTM network we use RNNLIB a recurrent neural network library \cite{rnnlib}.

\section{Dataset}
Indic scripts have huge number of classes due to the presence of basic characters, vowel modifiers and conjuncts. These conjuncts and vowel modifiers are composed of more than one unicode, thus learning the alignment of unicodes becomes important. This necisitates the usage of unicode re-ordering or post-processing schemes but LSTM using CTC output layer is able to learn the sequence alignment. Recognition of Oriya characters is very challenging due to the presence of large number of classes and highly similar shapes of basic characters. Pages are scanned from several books with different fonts at 300 dpi resolution and are binarized using Sauvola binarization. The pages do not have any skew but are heavily degraded as the books are very old. The foreground text has significant intereference from the background text due to thin pages. Raw binarized image pixels are used as input features by the network.


\section{Results}

For end to end recognition, different segmentation algorithms were used. In a traditional OCR workflow, the recognition accuracy suffers due to the presence of segmentation errors either at line /word/character level. The proposed framework gives us the freedom to choose from alternate segmentation hypothesis. The segmentation algorithms used as alternate hypothesis are complimentary in nature. 
As shown in table 1, individually Interval tree based segmentation(IT) performs worst in comparison with Geometric profiling and Hough transform. But by using all three as different branches for alternate segmentation, we observe better results in terms of both character and word recognition accuracies. In this paper we do not aim to bring the best segmentation algorithm, rather intend to use different segmentation pathways in order to improve recognition. Most errors arose from line segmentation, although we have observed some merged and broken words from the word segmentation routines. In case of interval tree based segmentation, the upper and lower zone characters got separated from their line and appeared as a different line thus increasing the number of lines. In case of Hough transform we use certain heuristics to restrain the line height to a average line height calculated over the training pages. Geometric profiling based methods worked better in comparison to other algorithms considered for line segmentation but it has immense usage of heuristics and spatial constraints. All these errors make it difficult to compare words with other words from different hypothesis since the number of lines and words out of each hypothesis is different. To solve this problem, we use a neighborhood search while traversing across a sub-string in search of valid of n-grams. If there is a mismatch in the different pathways, mostly this problem gets cascaded in successive words to generate more errors. By performing a search with preceding and succeeding words, we have been able to successfully solve such errors. The incorporation of context search benefited the framework as explained by an illustration in 1st row of figure 3. In this case we had two words which were segmented as a single word by IT and Hough Transform but as two different words by profiling. Due to the use of context search we could find the corresponding word in the next node and thus recognition is correct. We observed that mostly the sequences were picked from geometric profiling but in case of words which did not have upper or lower zone characters, interval tree based segmentation complimented the other hypotheses and resulted in correct sequences. In the 2nd row of figure 3, the word image is not discernable and is also misrecognized as a similar modifier with the exact shape exists. This had been correctly recognized since IT performed better on such middle zone characters. If a sub-string does not match with an n-gram, an error penalty is imposed and matching would continue for each word across all pathways. At each node the word with least error would be picked as the best word. 

\begin{figure}[h]
\centering
\includegraphics[width = 7 cm,height = 3cm]{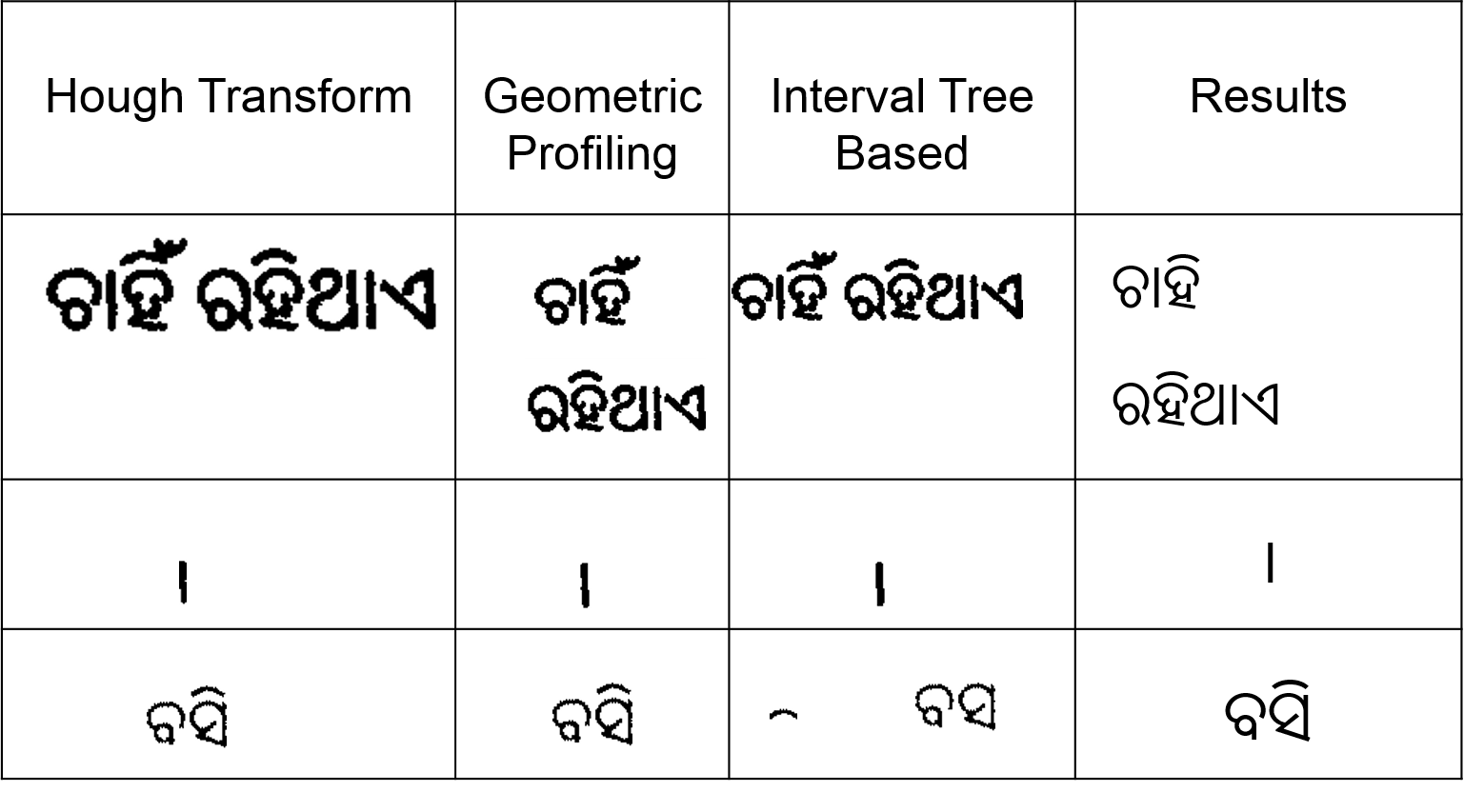}
\centering
\caption{Segments from individual segmentation algorithms and results from proposed framework}
\label{fig 3}
\end{figure} 

Due to the use of alternate hypothesis in finding the best word, this framework is able to take care of insertions and deletions which mainly arise out of the recognizer. When the substring is valid according to the n-grams but there is a substitution of any one or more than one primitives then this framework is unable to detect it. As we are working with full unsegmented words, the presence of a valid n-gram does not necessarily enforce correct recognition as there might exist a similar n-gram with some substitution which is also valid. Figure 4 shows parts of page images where there is a huge line segmentation error(highlighted in red boxes). This occurs due to the presence of lower zone modifiers in the upper line and upper zone modifiers in the lower line, which decreases the interline gap. In such cases the alignment of words also gets distorted due to change in number of lines and words in different hypotheses. Our framework consistently solves such issues due to the use of neighborhood during best first search and proves to be extremely effective. 

\begin{figure*}[t]
\subfigure[]{\includegraphics[width = 9cm, height = 2.5cm]{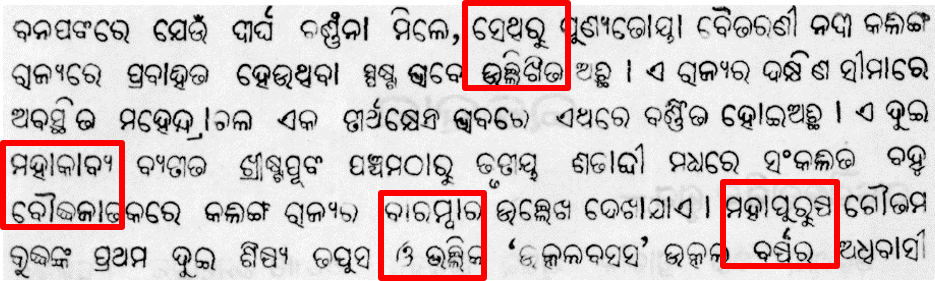}}
\subfigure[]{\includegraphics[width = 9cm, height = 2.5cm]{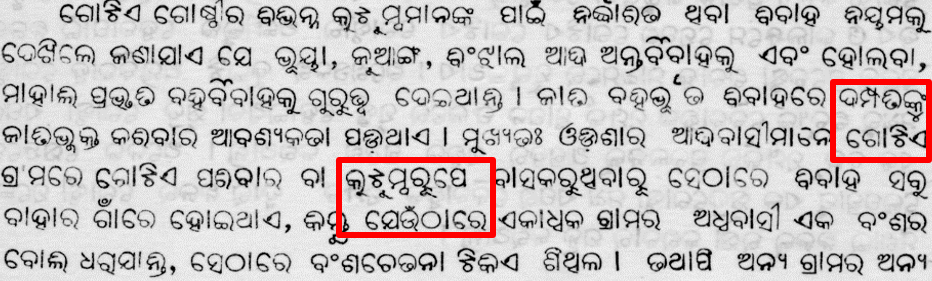}}
\caption{Figure show parts of pages where lines get merged due to lower zone of upper line and upper zone characters of lower line} 
\label{fig4}
\end{figure*}

We test the pages obtained using the proposed framework and calculate character and word recognition error which is given below in table 1. Due to the multiple hypotheses and verification framework we are able to obtain very high word recognition error. 
\begin{table}[h]
\caption{Test Results}
\label{table 1}
\centering
\begin{tabular}{|c | c| c|}
\hline
Method & Label Error(\%) & Word error rate\\
\hline
Geometric profiling & 14.10 & 16.301\\
\hline
Interval tree based Segmentation & 30.22 & 35.06\\
\hline
Hough Transform & 22.24 & 28.49\\
\hline
Proposed Framework & 8.64 & 10.64\\
\hline 
\end{tabular}
\end{table}

\section{Conclusion}

This paper proposes a text recognition framework which uses multiple segmentation algorithms as different hypotheses generators, recognizes each segment using a deep BLSTM network and verifies the performance of the deep neural network with a learned language model. In this work we segment words from a page using different segmentation routines and the best word is selected using best-first search over a spatial neighborhood to avoid alignment issues. The proposed framework obtained very high word recognition rate due to the use of alternate segmentation and verification using n-grams which helped filtering recognition errors. This framework is highly suitable for degraded documents wherein segmentation algorithms are the main causes of error. This framework is very effective in case of insertion and deletion errors introduced by the recognizer. If the segmentation algorithms considered are complimentary, the recognition error of the hybrid can be expected to be much less than the best segmentation framework. Deep BLSTM helps in recognizing sequences of words and also learns the alignment of unicodes, which is a challenge in Indic scripts. This work could be extended to recognize and verify longer text sequences.

\section*{Acknowledgment}
The authors would like to thank Dr. Alex Graves for his constant help and support throughout the work. The authors would like to thank Ministry of Communication and Information Technology, Government of India for the funding under the project titled Development of Robust Document Analysis and Recognition System for Printed Indian Scripts.



%
\bibliographystyle{IEEEtran}   			
\bibliography{final}  

\begin{thebibliography}{10}
\providecommand{\url}[1]{#1}
\csname url@samestyle\endcsname
\providecommand{\newblock}{\relax}
\providecommand{\bibinfo}[2]{#2}
\providecommand{\BIBentrySTDinterwordspacing}{\spaceskip=0pt\relax}
\providecommand{\BIBentryALTinterwordstretchfactor}{4}
\providecommand{\BIBentryALTinterwordspacing}{\spaceskip=\fontdimen2\font plus
\BIBentryALTinterwordstretchfactor\fontdimen3\font minus
  \fontdimen4\font\relax}
\providecommand{\BIBforeignlanguage}[2]{{%
\expandafter\ifx\csname l@#1\endcsname\relax
\typeout{** WARNING: IEEEtran.bst: No hyphenation pattern has been}%
\typeout{** loaded for the language `#1'. Using the pattern for}%
\typeout{** the default language instead.}%
\else
\language=\csname l@#1\endcsname
\fi
#2}}
\providecommand{\BIBdecl}{\relax}
\BIBdecl

\bibitem{my3rd}
A.~Ray, S.~Rajeswar, and S.~Chaudhury, ``Text recognition using deep blstm
  network,'' 2015.

\bibitem{Plamondon:2000:OOH:331097.331275}
R.~Plamondon and S.~N. Srihari, ``On-line and off-line handwriting recognition:
  A comprehensive survey,'' \emph{IEEE Trans. Pattern Anal. Mach. Intell.},
  vol.~22, no.~1, pp. 63--84, Jan. 2000.

\bibitem{graves2004biologically}
A.~Graves, D.~Eck, N.~Beringer, and J.~Schmidhuber, ``Biologically plausible
  speech recognition with lstm neural nets,'' in \emph{Biologically Inspired
  Approaches to Advanced Information Technology}.\hskip 1em plus 0.5em minus
  0.4em\relax Springer, 2004, pp. 127--136.

\bibitem{graves2005framewise}
A.~Graves and J.~Schmidhuber, ``Framewise phoneme classification with
  bidirectional lstm and other neural network architectures,'' \emph{Neural
  Networks}, vol.~18, no.~5, pp. 602--610, 2005.

\bibitem{liwicki2007novel}
M.~Liwicki, A.~Graves, H.~Bunke, and J.~Schmidhuber, ``A novel approach to
  on-line handwriting recognition based on bidirectional long short-term memory
  networks,'' in \emph{Proc. 9th Int. Conf. on Document Analysis and
  Recognition}, vol.~1, 2007, pp. 367--371.

\bibitem{6854518}
R.~Brueckner and B.~Schulter, ``Social signal classification using deep blstm
  recurrent neural networks,'' in \emph{Acoustics, Speech and Signal Processing
  (ICASSP), 2014 IEEE International Conference on}, May 2014, pp. 4823--4827.

\bibitem{wollmer2010context}
M.~W{\"o}llmer, A.~Metallinou, F.~Eyben, B.~Schuller, and S.~S. Narayanan,
  ``Context-sensitive multimodal emotion recognition from speech and facial
  expression using bidirectional lstm modeling.'' in \emph{INTERSPEECH}, 2010,
  pp. 2362--2365.

\bibitem{GravesS08}
A.~Graves and J.~Schmidhuber, ``Offline handwriting recognition with
  multidimensional recurrent neural networks.'' in \emph{NIPS}.\hskip 1em plus
  0.5em minus 0.4em\relax Curran Associates, Inc., 2009, pp. 545--552.

\bibitem{NIPS2007_3213}
A.~Graves, M.~Liwicki, H.~Bunke, J.~Schmidhuber, and S.~Fern\'{a}ndez,
  ``Unconstrained on-line handwriting recognition with recurrent neural
  networks,'' in \emph{Advances in Neural Information Processing Systems
  20}.\hskip 1em plus 0.5em minus 0.4em\relax Curran Associates, Inc., 2008,
  pp. 577--584.

\bibitem{Ul-Hasan:2013:OPU:2549400.2549465}
A.~Ul-Hasan, S.~B. Ahmed, F.~Rashid, F.~Shafait, and T.~M. Breuel, ``Offline
  printed urdu nastaleeq script recognition with bidirectional lstm networks,''
  in \emph{Proceedings of the 2013 12th International Conference on Document
  Analysis and Recognition}, ser. ICDAR '13.\hskip 1em plus 0.5em minus
  0.4em\relax Washington, DC, USA: IEEE Computer Society, 2013, pp. 1061--1065.

\bibitem{Breuel:2013:HOP:2549400.2549524}
T.~M. Breuel, A.~Ul-Hasan, M.~A. Al-Azawi, and F.~Shafait, ``High-performance
  ocr for printed english and fraktur using lstm networks,'' in
  \emph{Proceedings of the 2013 12th International Conference on Document
  Analysis and Recognition}, ser. ICDAR '13.\hskip 1em plus 0.5em minus
  0.4em\relax Washington, DC, USA: IEEE Computer Society, 2013, pp. 683--687.

\bibitem{Rashid:2013}
S.~F. Rashid, M.-P. Schambach, J.~Rottland, and S.~von~der N\"{u}ll, ``Low
  resolution arabic recognition with multidimensional recurrent neural
  networks,'' in \emph{Proceedings of the 4th International Workshop on
  Multilingual OCR}, ser. MOCR '13.\hskip 1em plus 0.5em minus 0.4em\relax New
  York, NY, USA: ACM, 2013, pp. 6:1--6:5.

\bibitem{834}
``Feature design for offline arabic handwriting recognition: handcrafted vs
  automated?'' in \emph{12th International Conference on Document Analysis and
  Recognition (ICDAR {\textquoteright}13)}, 2013.

\bibitem{hinton2006fast}
G.~Hinton, S.~Osindero, and Y.-W. Teh, ``A fast learning algorithm for deep
  belief nets,'' \emph{Neural computation}, vol.~18, no.~7, pp. 1527--1554,
  2006.

\bibitem{venu:09}
V.~Govindaraju and S.~Setlur, \emph{Guide to OCR for Indic Scripts}.\hskip 1em
  plus 0.5em minus 0.4em\relax Springer, 2009.

\bibitem{953897}
B.~Chaudhuri, U.~Pal, and M.~Mitra, ``Automatic recognition of printed oriya
  script,'' in \emph{Document Analysis and Recognition, 2001. Proceedings.
  Sixth International Conference on}, 2001, pp. 795--799.

\bibitem{DBLP:conf/icfhr/FinkVBPC10}
G.~A. Fink, S.~Vajda, U.~Bhattacharya, S.~K. Parui, and B.~B. Chaudhuri,
  ``Online bangla word recognition using sub-stroke level features and hidden
  markov models,'' in \emph{International Conference on Frontiers in
  Handwriting Recognition, {ICFHR} 2010, Kolkata, India, 16-18 November 2010},
  2010, pp. 393--398.

\bibitem{DBLP:conf/icpr/ParuiGBC08}
S.~K. Parui, K.~Guin, U.~Bhattacharya, and B.~B. Chaudhuri, ``Online
  handwritten bangla character recognition using {HMM},'' in \emph{19th
  International Conference on Pattern Recognition {(ICPR} 2008), December 8-11,
  2008, Tampa, Florida, {USA}}, 2008, pp. 1--4.

\bibitem{NaveenICPR}
N.~Sankaran and C.~V. Jawahar, ``Recognition of printed devanagari text using
  blstm neural network,'' in \emph{ICPR'12}, 2012, pp. 322--325.

\bibitem{NaveenICDAR}
N.~Sankaran, A.~Neelappa, and C.~Jawahar, ``Devanagari text recognition: A
  transcription based formulation,'' in \emph{Document Analysis and Recognition
  (ICDAR), 2013 12th International Conference on}, Aug 2013, pp. 678--682.

\bibitem{NaveenDAS}
S.~Dutta, N.~Sankaran, K.~P. Sankar, and C.~V. Jawahar, ``Robust recognition of
  degraded documents using character n-grams,'' in \emph{Document Analysis
  Systems'12}, 2012, pp. 130--134.

\bibitem{Graves:2009:NCS:1525650.1525782}
A.~Graves, M.~Liwicki, S.~Fern\'{a}ndez, R.~Bertolami, H.~Bunke, and
  J.~Schmidhuber, ``A novel connectionist system for unconstrained handwriting
  recognition,'' \emph{IEEE Trans. Pattern Anal. Mach. Intell.}, vol.~31,
  no.~5, pp. 855--868.

\bibitem{Graves:2006:CTC:1143844.1143891}
A.~Graves, S.~Fern\'{a}ndez, F.~Gomez, and J.~Schmidhuber, ``Connectionist
  temporal classification: Labelling unsegmented sequence data with recurrent
  neural networks,'' in \emph{Proceedings of the 23rd International Conference
  on Machine Learning}, 2006, pp. 369--376.

\bibitem{Hochreiter:1997}
S.~Hochreiter and J.~Schmidhuber, ``Long short-term memory,'' \emph{Neural
  Comput.}, vol.~9, no.~8, pp. 1735--1780, Nov. 1997.

\bibitem{DBLP:conf/asru/GravesJM13}
A.~Graves, N.~Jaitly, and A.~rahman Mohamed, ``Hybrid speech recognition with
  deep bidirectional lstm,'' in \emph{ASRU}, 2013, pp. 273--278.

\bibitem{DBLP:journals/corr/abs-1303-5778}
A.~Graves, A.~rahman Mohamed, and G.~E. Hinton, ``Speech recognition with deep
  recurrent neural networks,'' \emph{CoRR}, vol. abs/1303.5778, 2013.

\bibitem{rnnlib}
A.~Graves, ``Rnnlib: A recurrent neural network library for sequence learning
  problems,'' http://sourceforge.net/projects/rnnl/.

\end{thebibliography}
%
%

\end{document}